\documentclass{article}
\usepackage{ijcai26}
\usepackage{acronym}

\usepackage{amsfonts}

\usepackage{color}
\usepackage{times}
\usepackage{soul}
\usepackage{url}
\usepackage[hidelinks]{hyperref}
\usepackage[utf8]{inputenc}
\usepackage[small]{caption}
\usepackage{graphicx}
\usepackage{amsmath}
\usepackage{amsthm}
\usepackage{multirow}
\usepackage{booktabs}
\usepackage{algorithm}
\usepackage{algorithmic}
\usepackage[switch]{lineno}

\title{IB-GRPO: Aligning LLM-based Learning Path Recommendation with Educational Objectives via Indicator-Based Group Relative Policy Optimization}

\author{
Shuai Wang\#$^{1}$
\and
Yaoming Yang\#$^1$\and
Bingdong Li* $^{1}$\and
Hao Hao$^1$\And
Aimin Zhou$^1$\\
\affiliations
$^1$East China Normal University
\emails
52285901025@stu.ecnu.edu.cn,
51275901039@stu.ecnu.edu.cn,
 bdli@cs.ecnu.edu.cn,
hhao@mail.ecnu.edu.cn,
amzhou@cs.ecnu.edu.cn
}

\begin{document}
\maketitle 

\begin{abstract}
Learning Path Recommendation (LPR) aims to generate personalized sequences of learning items that maximize long-term learning effect while respecting pedagogical principles and operational constraints. Although large language models (LLMs) offer rich semantic understanding for free-form recommendation, applying them to long-horizon LPR is challenging due to (i) misalignment with pedagogical objectives such as the Zone of Proximal Development (ZPD) under sparse, delayed feedback, (ii) scarce and costly expert demonstrations, and (iii) multi-objective interactions among learning effect, difficulty scheduling, length controllability, and trajectory diversity. To address these issues, we propose IB-GRPO (Indicator-Based Group Relative Policy Optimization), an indicator-guided alignment approach for LLM-based LPR. To mitigate data scarcity, we construct hybrid expert demonstrations via Genetic Algorithm search and teacher RL agents and warm-start the LLM with supervised fine-tuning. Building on this warm-start, we design a within-session ZPD alignment score for difficulty scheduling. IB-GRPO then uses the $I_{\epsilon+}$ dominance indicator to compute group-relative advantages over multiple objectives, avoiding manual scalarization and improving Pareto trade-offs. Experiments on ASSIST09 and Junyi using the KES simulator with a Qwen2.5-7B backbone show consistent improvements over representative RL and LLM baselines.

\end{abstract}

\section{Introduction}

Learning Path Recommendation (LPR) is a key component of adaptive learning and has been increasingly adopted in recent intelligent tutoring systems~\cite{zhang2024item}. Unlike one-size-fits-all instruction, LPR aims to generate personalized sequences of learning items (e.g., exercises) that bridge the gap between a student's current knowledge state and their learning goals.

Many recent approaches typically model LPR as a Markov Decision Process (MDP)~\cite{li2023graph,pateria2021hierarchical}, using Reinforcement Learning (RL) to optimize long-term rewards. However, traditional RL methods rely on ID-based representations. By treating exercises as simple IDs, they ignore the rich semantic information in the content, leading to rigid recommendations~\cite{lv2025real}.

Recently, Large Language Models (LLMs) have opened new possibilities for LPR by leveraging their powerful semantic understanding and world knowledge~\cite{lv2025real,lv2025genal}. Generative agents based on LLMs can effectively interpret exercise texts and simulate tutor-like interactions, mitigating the cold-start problem faced by ID-based methods. However, directly deploying off-the-shelf LLMs for this long-horizon planning task entails two fundamental challenges. \textbf{(1) The misalignment between generic pre-training and pedagogical objectives.}  LLMs are primarily trained for next-token prediction, which often leads to myopic decision-making—optimizing for immediate plausibility rather than maximizing a student's long-term learning effect~\cite{meincke2024beyond}. While recent agent-based~\cite{lv2025genal} methods utilize prompting or retrieval to guide the model, they lack explicit parameter updates to strictly enforce educational theories, such as the Zone of Proximal Development (ZPD)~\cite{shabani2010vygotsky}. Without fine-tuning on pedagogical objectives, LLMs struggle to appropriately calibrate difficulty levels to differences in students' abilities over a long sequence. Addressing this via RL is non-trivial due to the scarcity of high-quality supervision signals. Constructing learning paths that strictly adhere to ZPD is intricate, and random exploration in the vast combinatorial space of LLMs is inefficient, making the "cold start" for RL alignment particularly difficult. \textbf{(2) The complexity of multi-objective decision making.} Realistic LPR must optimize the delayed learning outcome while also respecting process-level pedagogical principles (e.g., ZPD-aligned difficulty scheduling), operational constraints (e.g., lesson length), and trajectory-level coverage (diversity). While ZPD alignment is generally consistent with learning improvement, it is not guaranteed by optimizing a sparse outcome signal alone; moreover, diversity and length constraints introduce explicit trade-offs with learning effect under a fixed budget. This naturally leads to a multi-objective formulation. Prevailing methods, including the recent LLM-based educational agent Pxplore~\cite{lim2025personalized}, largely rely on scalarization, compressing multi-dimensional rewards into a single scalar via linear weighting~\cite{machado2025dylam}. This approach is not only dependent on tedious manual tuning but also prone to "locking" policy preferences when objectives conflict, making it difficult to approximate the Pareto frontier~\cite{he2025pareto,song2025balancing}.

To address these challenges, we propose IB-GRPO, a two-stage alignment approach for LLM-based LPR that integrates pedagogical signals with practical generation constraints. Our main contributions are:

\begin{itemize}
    \item we design a hybrid expert data synthesis pipeline that combines Genetic Algorithm (GA) search with teacher RL policies, producing diverse and high-quality learning-path demonstrations for supervised warm-start.
    
    \item We formalize ZPD as a reward for difficulty alignment and propose Indicator-Based Group Relative Policy Optimization (IB-GRPO) to handle multi-objective trade-offs. IB-GRPO uses the $I_{\epsilon+}$ dominance indicator to compute group-relative advantages over vector rewards, improving Pareto trade-offs.

    \item Extensive experiments demonstrate consistent improvements in learning effect compared to RL and LLM baselines, while maintaining ZPD compliance.
\end{itemize}

\section{Related Work}

\subsection{Learning Path Recommendation (LPR)}

Prior LPR research spans several lines. Knowledge-modeling based methods leverage Cognitive Diagnosis Models (CDMs), knowledge tracing, or Knowledge Graphs to estimate learner mastery and recommend subsequent items~\cite{wang2022neuralcd}. Although effective for structured assessment, they often depend on hand-crafted assumptions or heuristics and do not explicitly optimize long-horizon outcomes in sequential recommendations. More recent work formulates LPR as sequential decision making and adopts Deep Reinforcement Learning (DRL) to optimize long-term learning effects. For instance, CSEAL~\cite{liu2019exploiting} employs an actor-critic framework integrated with a cognitive navigation mechanism, while RLTutor~\cite{dwivedi2018learning} combines model-based RL with the DAS3H tracing model. Recent RL-based LPR further improves planning robustness by injecting richer structural priors, such as explicitly modeling item difficulty via hierarchical graphs and difficulty-driven hierarchical RL~\cite{zhang2024item}. Some methods also leverage privileged signals, for example through graph-enhanced hierarchical planning with teacher-state distillation~\cite{li2024privileged}. However, these methods critically rely on ID-based representations.

\subsection{Large Language Models in Education}
Leveraging their powerful context-awareness and reasoning capabilities, LLMs have catalyzed a new paradigm of educational agents~\cite{wang2024survey}. Current exploration focuses on two main avenues: (1) Educational Simulation and Tutoring Agents. Systems like Agent4Edu~\cite{gao2025agent4edu} and EduAgent~\cite{xu2024eduagent} utilize LLMs to simulate learner interactions with educational content (e.g., exercises, videos) and evaluate learning outcomes. (2) Knowledge-based Recommendation. Some studies exploit the conceptual knowledge of LLMs for next-concept recommendation~\cite{li2024learning}. Despite significant progress, existing Agent-based methods predominantly rely on prompt engineering, lacking parameter update mechanisms to optimize long-range pedagogical objectives. Recently, Pxplore~\cite{lim2025personalized} advanced this by employing a ``SFT + RL'' paradigm to align LLMs with educational goals. However, it still resorts to scalarized rewards with empirical weights, which struggles to dynamically balance conflicting objectives, and its supervised initialization incurs high costs due to dependence on GPT-4o-based annotations.

\subsection{Multi-Objective RL and Alignment}
Aligning LLMs often requires managing trade-offs across multiple objectives. Standard Reinforcement Learning from Human Feedback (RLHF) methods (e.g., PPO~\cite{ppo}, DPO~\cite{dpo}) typically scalarize vector rewards, obscuring trade-offs. Multi-Objective Reinforcement Learning attempts to solve this by directly optimizing the Pareto frontier. Methods like DyLam~\cite{machado2025dylam} dynamically adjust weights, while recent advances such as HVO~\cite{song2025balancing} and MO-GRPO~\cite{ichihara2025mo} utilize Hypervolume (HV) contribution as a reward signal. GDPO~\cite{liu2026gdpo} proposes a reward-decoupled normalization strategy to prevent advantage collapse in multi-objective settings. However, LPR involves a densely clustered solution space due to high path homogeneity. In such scenarios, volume-based rewards (e.g., HVO) suffer from vanishing gradients as marginal hypervolume contributions become negligible. To address this, we propose IB-GRPO, utilizing the $I_{\epsilon+}$ indicator to capture pairwise dominance. Unlike volume metrics, this indicator extracts granular gradient signals even among highly similar candidates, ensuring effective optimization towards the Pareto frontier.

\section{Problem Formulation}
\label{sec3}
In this section, we formally define the Multi-Objective Personalized Learning Path Recommendation (MOLPR) task. We formulate the recommendation process as a sequential decision-making problem, and then specify four optimization objectives that constitute our multi-objective formulation and the reward vector optimized in our framework.

\subsection{Notation and Task Definition}
Let $\mathcal{C} = \{c_1, c_2, \dots, c_{|\mathcal{C}|}\}$ denote the set of knowledge concepts. A learner's interaction history is represented as $H = \{(c_1, y_1), \dots, (c_k, y_k)\}$, where $c_i \in \mathcal{C}$ is the practiced concept and $y_i \in \{0, 1\}$ denotes the response correctness. We use $\pi$ to denote a learning path (trajectory), and $\mu_{\theta}$ to denote the recommendation policy parameterized by $\theta$. In this work, we model the learner's proficiency as a stationary latent variable $a \in \mathbb{R}$.

\subsection{Optimization Objectives}
We formulate four key objectives to guide the policy optimization:

\subsubsection{Objective 1: Learning Effect Maximization ($E_p$)}
The primary goal of education is to enhance learner proficiency. We quantify the learning effect as the normalized improvement in the learner's test scores after completing the recommended path:
\begin{equation}
    E_p(\pi) = \frac{E_{e} - E_{s}}{E_{sup} - E_{s}}
\end{equation}
where $E_{e}$ and $E_{s}$ represent the learner's scores on a comprehensive post-test and pre-test, respectively, and $E_{sup}$ is the maximum possible score for the test. This metric serves as a direct outcome-oriented indicator.

\subsubsection{Objective 2: Optimal Challenge via ZPD ($S_{ZPD}$)}

Drawing on the ZPD theory~\cite{shabani2010vygotsky}, effective learning occurs when the task difficulty aligns with the learner's proficiency~\cite{vermeiren2025psychometrics}. Let $d(\pi_t)$ denote the difficulty of a recommended concept, which is defined by the error rate of all students. Instead of using a heuristic rule, we define $z(a)$ as the center of the empirical optimal challenge distribution, which is estimated from offline trajectories to capture the difficulty range that historically yields maximum learning effects for a given proficiency $a$. In practice, we estimate $z(a)$ from offline trajectories by aggregating the difficulties of concepts appearing in high-outcome trajectories with $E_p > 0.9$ for students with proficiency $a$, and use a Gaussian-shaped kernel to softly measure how close $d(\pi_t)$ is to this data-driven center. The ZPD alignment score measures the consistency between the recommendation and this data-driven reference:
\begin{equation}
    S_{ZPD}(\pi) = \frac{1}{L} \sum_{t=1}^{L} \exp\left(-\frac{(d(\pi_{t}) - z(a))^2}{2\sigma^2}\right)
\end{equation}
Here, $\sigma$ controls the spread of the optimal difficulty distribution. This Gaussian formulation defines ZPD as a soft region centered on $z(a)$, encouraging the policy to adhere to the empirically observed optimal learning zone, ensuring pedagogical adaptability.

\subsubsection{Objective 3: Length Constraint Satisfaction ($R_{Len}$)}
To ensure the recommended path adheres to the lesson planning constraints, we impose a constraint on the path length. Let $L_{target}$ be the desired length and $|\pi|$ be the actual length of the generated path. We define the length deviation as $\Delta = \left| |\pi| - L_{target}\right|$.
The length reward function is defined as:
\begin{equation}
    R_{Len}(\pi) = 
    \begin{cases} 
    1.0, & \text{if } \Delta \le \tau \\
    -\lambda (\Delta - \tau), & \text{if } \Delta > \tau
    \end{cases}
\end{equation}
where $\tau$ is the tolerance threshold and $\lambda$ is a penalty coefficient. This design applies a linear penalty only when the length deviation exceeds the strictly allowed margin.

\subsubsection{Objective 4: Diversity ($D_{Div}$)}
To prevent the policy from converging to repetitive loops or homogeneous patterns, we explicitly encourage diversity within each generation batch.
Each path $\pi$ is represented as an ordered sequence of concept IDs, and we measure similarity using Jaccard similarity over its $n$-gram feature set.
Formally, let $G_n(\pi)$ denote the set of all contiguous $n$-grams extracted from the concept sequence of $\pi$. We define
\begin{equation}
\mathrm{Sim}_{\mathrm{Jaccard}}(\pi,\pi') \;=\; \frac{|G_n(\pi)\cap G_n(\pi')|}{|G_n(\pi)\cup G_n(\pi')|}
\end{equation}
The diversity reward for $\pi$ with respect to a comparison set $\mathcal{B}$ is
\begin{equation}
\begin{split}
D_{{Div}}(\pi;\mathcal{B}) &= 1 - \frac{1}{|\mathcal{B}|-1} \\
&\quad \sum_{\pi'\in\mathcal{B},\,\pi'\neq\pi}\mathrm{Sim}_{\mathrm{Jaccard}}(\pi,\pi')
\end{split}
\end{equation}
Here, $\mathcal{B}$ is taken as the set of samples that share the same prompt within the batch (prompt-grouped diversity). This objective drives the agent to explore the solution space by penalizing paths with highly overlapping $n$-gram patterns compared to others in the same group.
\subsection{Overall Optimization Goal}
Finally, the MOLPR task is formulated as learning a stochastic policy $\mu_\theta$ that induces a distribution over learning paths $\pi \in \Pi$. In our setting, $E_p$ serves as the primary outcome objective, while $S_{ZPD}$ provides an auxiliary difficulty-alignment signal to support stable learning effect. $R_{Len}$ and $D_{Div}$ address issues unique to LLM-based free-form path generation, namely length controllability and repetitive recommendations. For each path $\pi$, we define a vector-valued objective:
\begin{equation}
    \mathbf{J}(\pi) = [E_{p}(\pi), S_{ZPD}(\pi), R_{Len}(\pi), D_{Div}(\pi)]^\top.
\end{equation}
A learning path $\pi^*$ is Pareto optimal if there exists no other path $\pi' \in \Pi$ such that $\mathbf{J}(\pi') \ge \mathbf{J}(\pi^*)$ (component-wise) and $\mathbf{J}(\pi') \neq \mathbf{J}(\pi^*)$.
The Pareto frontier is the set of all Pareto-optimal paths in $\Pi$.
Our goal is to learn $\mu_\theta$ such that the non-dominated paths sampled from $\mu_\theta$ approximate this Pareto frontier. Rather than scalarizing these objectives using fixed weights, we optimize the vector objective by leveraging intra-group dominance comparisons, enabling the policy to discover superior trade-offs.

\section{Methodology}
\label{sec:methodology}

\begin{figure*}[t]
    \centering
    \includegraphics[width=0.95\linewidth]{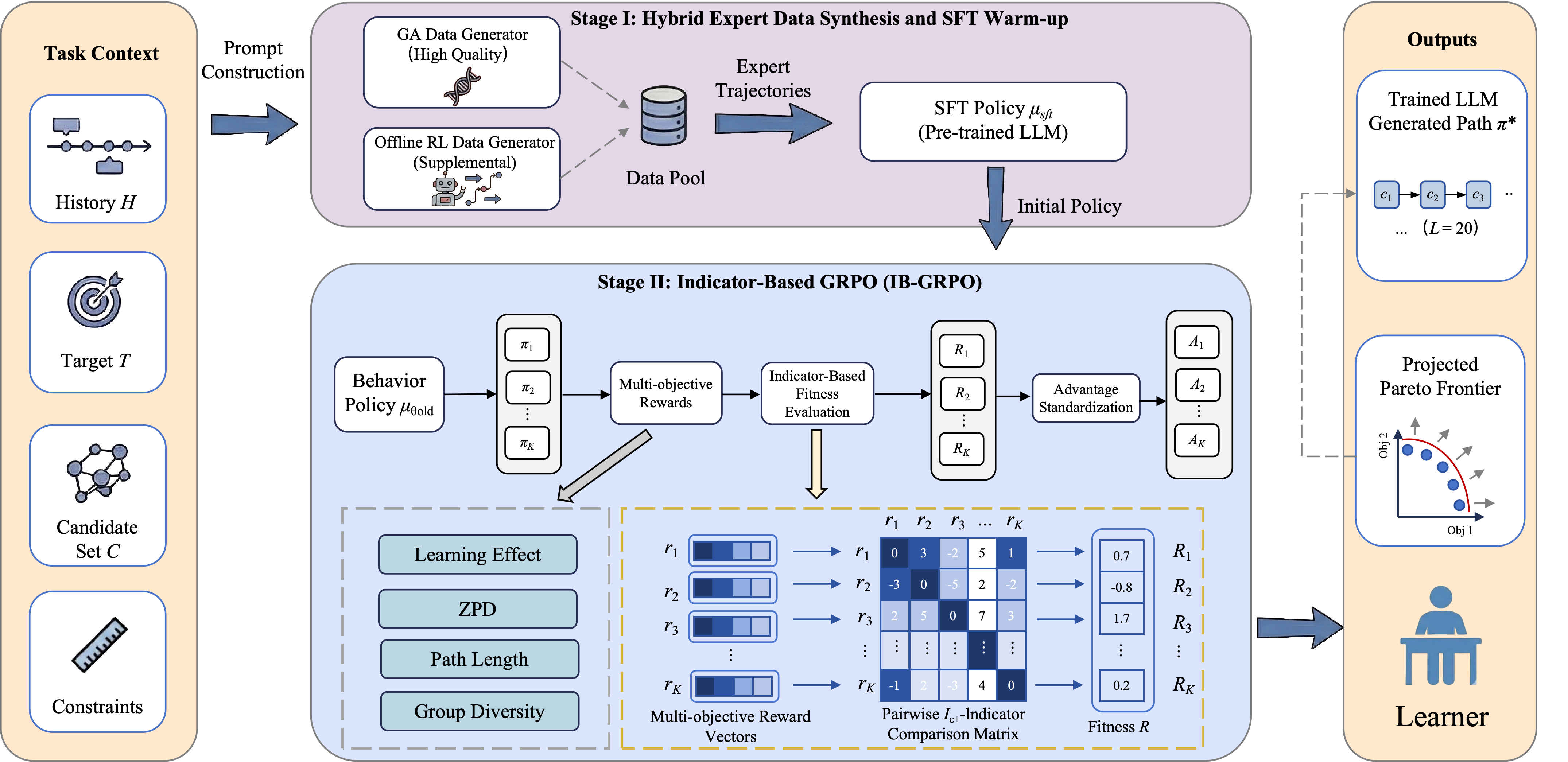} 
    \caption{Overall framework of IB-GRPO. \textbf{Stage I (top):} Hybrid expert data synthesis (GA search + offline RL teachers) produces demonstration paths for SFT, yielding a model $\mu_{\mathrm{sft}}$. \textbf{Stage II (bottom):} Starting from the SFT model, we sample $K$ candidate paths per prompt, compute multi-objective rewards, and derive group-relative advantages via the $I_{\epsilon+}$-based fitness to update the policy toward the Pareto frontier.}
    \label{fig:framework}
\end{figure*}

In this section, we introduce IB-GRPO (Figure~\ref{fig:framework}) to align LLM-based LPR with multi-objective pedagogical goals via two stages: hybrid expert SFT warm-up (GA search + offline RL teachers) yielding $\mu_{\mathrm{sft}}$, followed by IB-GRPO policy optimization with $I_{\epsilon+}$-based group-relative advantages over vector rewards to improve Pareto trade-offs without manual weight tuning.

\subsection{Data Synthesis via Hybrid Experts}
A critical prerequisite for on-policy Group Relative Policy Optimization (GRPO) is that the sampled group must exhibit sufficient diversity and quality trade-offs to compute meaningful advantages. Initializing with standard RL policies often leads to mode collapse, while random initialization yields low-quality noise. Therefore, Stage 1 is not merely for warm-up, but to construct an initial Pareto landscape. By distilling the global search capability of GA and the local exploitation of RL, we ensure the initial policy $\mu_{\mathrm{sft}}$ covers the efficiency-diversity trade-off front, providing a rich contrastive signal for the subsequent IB-GRPO stage.

\subsubsection{Source A: Search-based Expert (GA)}
For a given student state, we formulate the path recommendation as a discrete combinatorial optimization problem. Leveraging the global search capability of GA~\cite{katoch2021review}, GA search tends to produce diverse and reasonably high-quality paths, but its search efficiency is limited for consistently reaching high-score regions; in contrast, purely RL-based generation is prone to mode collapse and yields low diversity. As shown in RQ2, combining GA with teacher RL trajectories yields a more favorable $E_p$--Diversity trade-off than either source alone, resulting in a stronger warm-start distribution for SFT.

\begin{itemize}
    \item \textbf{Encoding:} We encode a learning path as a chromosome $\pi=[\pi_{1}, \pi_{2}, ..., \pi_{L}]$, where each gene corresponds to a recommended knowledge concept.
    \item \textbf{Evolution:} Through tournament selection, single-point crossover, and mutation operators, the GA evolves a population of candidate paths. This process excels at generating diverse samples that satisfy learning $E_{p}$, providing high-quality solutions that are often hard to reach via simple greedy policies.
\end{itemize}

\subsubsection{Source B: Policy-based Expert (RL Agents)}
While GA offers strong exploration, it incurs high computational costs and may converge slowly on complex student states. To ensure data coverage and robustness, we incorporate pre-trained traditional RL models (e.g., CSEAL~\cite{liu2019exploiting}, GEPKSD~\cite{li2024privileged}) as "Teacher Agents".
\begin{itemize}
    \item \textbf{Complementary Generation:} For states where GA fails to find satisfactory solutions efficiently, we query these Teacher Agents to generate recommended paths.
    \item \textbf{Structural Prior:} Although these traditional RL paths may lack deep semantic nuance, they possess strong structural integrity. They serve as essential supplementary data to stabilize the training distribution.
\end{itemize}

\subsubsection{Supervised Warm-up via Behavior Cloning}
We aggregate the high-quality paths generated by both GA and RL experts into a unified dataset $\mathcal{D}_{sft}=\{(s, \pi_{expert})\}$. Here, $s$ denotes the student-specific recommendation context used to construct the prompt, summarizing the interaction history $H$ together with the path-length constraint and other recommendation requirements. The Supervised Fine-Tuning (SFT) phase functions as a Behavior Cloning (BC) process, aiming to distill the planning capabilities of these experts into the LLM:
\begin{equation}
\begin{split}
    \mathcal{L}_{SFT}(\theta) &= -\mathbb{E}_{(s, \pi_{\mathrm{expert}}) \sim \mathcal{D}_{\mathrm{sft}}} \\
    &\qquad \left[ \sum_{t=1}^{L} \log \mu_{\theta}(\pi_{t} \mid s, \pi_{<t}) \right]
\end{split}
\end{equation}

Through this warm-up phase, the LLM acquires the fundamental paradigm of sequential path planning and domain-specific constraints. This effectively solves the policy cold-start problem, providing a solid foundation for the subsequent IB-GRPO training.

\subsection{Indicator-Based GRPO}

Although the SFT phase provides a foundational policy, it remains constrained by the static distribution of offline data and lacks the mechanism to explicitly handle scheduling ZPD adaptation and trade-offs across objectives. To address these limitations, we propose Indicator-Based Group Relative Policy Optimization (IB-GRPO).

\subsubsection{Vectorized Reward Modeling}
For a given student state $s$, each complete learning path generated by the model receives an $M$-dimensional reward vector ($M=4$ in our setting):
\begin{equation}
    r(s, \pi) = [{E_p}, S_{ZPD}, R_{Len}, D_{Div}]^\top
\end{equation}
where $S_{ZPD}$ is calculated based on the ZPD matching function defined in Section~\ref{sec3}.

\subsubsection{Group Sampling and Advantage Estimation}
For each input prompt, we sample a group of $K$ output paths $G=\{\pi_1,\ldots,\pi_K\}$ and compute a vector reward $r_i \in \mathbb{R}^M$ for each path $\pi_i$. To evaluate the quality of paths without reducing them to a scalar via linear weighting, we employ the $I_{\epsilon+}$-indicator~\cite{zitzler2004indicator} to measure the dominance relationship between any two paths $\pi_{j}$ and $\pi_{i}$ within the group. It is defined as:
\begin{equation}
    I_{\epsilon+}(r_{j}, r_{i}) = \max_{m \in \{1, ..., M\}} (r_{i,m} - r_{j,m})
\end{equation}
Here, $r_{i,m}$ denotes the reward value of path $\pi_{i}$ on the $m$-th objective. Intuitively, $I_{\epsilon+}$ quantifies the minimum amount by which $r_{j}$ must be improved in all dimensions to weakly dominate $r_{i}$.

Based on this indicator, we define the Pareto Fitness $R_{i}$ of path $\pi_{i}$ within group $G$ as the sum of exponential losses, inspired by Indicator-Based Evolutionary Algorithm (IBEA)~\cite{zitzler2004indicator}:
\begin{equation}
    R_{i} = \sum_{\pi_{j} \in G, j \neq i} - \exp \left( - \frac{I_{\epsilon+}(r_{j}, r_{i})}{\kappa} \right)
\end{equation}
where $\kappa > 0$ is a scaling factor regulating selection pressure. Unlike weighted sum methods which imply linear trade-offs, this indicator-based approach allows the policy to capture Pareto frontiers. If $\pi_{i}$ is strictly dominated by other paths in the group (yielding large negative $I_{\epsilon+}$ values), its fitness $R_{i}$ is severely penalized. Conversely, paths residing near the local Pareto frontier receive higher fitness scores. Note that the computational overhead of this pairwise comparison is $O(K^2)$, which is negligible given the typically small group size $K$ in GRPO~\cite{shao2024deepseekmath}.

To derive the gradient signal for policy updates, we compute the Group Relative Advantage by standardizing the fitness $R_{i}$ within group $G$:
\begin{equation}
    A_{i} = \frac{R_{i} - \text{Mean}(\{R_{k}\}_{k=1}^{K})}{\text{Std}(\{R_{k}\}_{k=1}^{K}) + \epsilon}
\end{equation}
Through standardization, paths performing above the group's average Pareto efficiency obtain a positive advantage $A_{i} > 0$, thereby reinforcing the policy. This enables IB-GRPO to adaptively optimize towards the Pareto frontier without manual weight tuning.

\subsubsection{Policy Optimization Objective}
Leveraging the computed intra-group multi-objective advantage $A_{i}$, we maximize the following objective function to update the policy parameters $\theta$:
\begin{equation}
\mathcal{L}_{IB}(\theta)
=
\mathbb{E}\left[
\frac{1}{K}\sum_{i=1}^{K}
\frac{\mu_{\theta}(\pi_{i}\mid s)}{\mu_{\theta_{old}}(\pi_{i}\mid s)} A_{i}
\right]
\end{equation}

where $\mu_{\theta_{old}}$ represents the policy from which the group samples were collected. In implementation, we apply a decoupled asymmetric clipping mechanism to the importance ratio $\mu_{\theta}(\pi_{i}\mid s)/\mu_{\theta_{old}}(\pi_{i}\mid s)$ to stabilize updates, using clipping bounds $[1-\epsilon_{\text{low}},\,1+\epsilon_{\text{high}}]$ with $\epsilon_{\text{low}}=0.2$ and $\epsilon_{\text{high}}=0.28$.

\section{Experiments}
\label{sec:experiments}

In this section, we address the following research questions through a series of experiments:
\begin{itemize}
    \item \textbf{RQ1:} How does IB-GRPO perform compared to state-of-the-art RL and LLM baselines across LPR tasks of varying lengths?
    \item \textbf{RQ2:} Can the hybrid offline data generation mechanism (combining GA and RL) effectively expand the Pareto frontier?
    \item \textbf{RQ3:} What are the individual contributions of the ZPD reward and the indicator-based advantage estimation module to the overall performance?
    \item \textbf{RQ4:} Do the learning path trajectories generated by the model align with students' cognitive development patterns?
\end{itemize}

\subsection{Experimental Setup}

\subsubsection{Datasets and Student Simulator}
We conduct experiments on two widely used public educational datasets: ASSIST09\footnote{https://sites.google.com/site/assistmentsdata/home}
 and Junyi\footnote{https://pslcdatashop.web.cmu.edu}
, both containing large-scale student interaction logs (see Table~\ref{t:data} for statistics). Since offline logs cannot provide feedback for counterfactual paths generated by a policy, we use the Knowledge Evolution Simulator (KES)~\cite{huang2019exploring} as the interaction environment. KES is a common LPR simulator that employs a Deep Knowledge Tracing (DKT) model to simulate learners' knowledge states. Following the data partitioning protocol in~\cite{li2024privileged}, we construct training splits to fit the DKT-based simulator, ensuring it captures the learning patterns in the data. At inference time, the simulator interacts with the recommendation policy to provide feedback for evaluating long-horizon learning effects.

\begin{table*}[!h]
\centering
\caption{Performance comparison of different algorithms across datasets and path lengths $L\in\{5,10,20\}$ (Metric: Learning Effect $E_p$, higher is better).}
\label{tab1}
\small
\setlength{\tabcolsep}{5pt}
\begin{tabular}{llccccccccc}
\toprule
\multirow{2}{*}{Dataset} & \multirow{2}{*}{length}
& \multicolumn{1}{c}{Non-RL Methods}
& \multicolumn{5}{c}{RL or RL-based Methods}
& \multicolumn{3}{c}{LLM-based Methods} \\
\cmidrule(lr){3-3}\cmidrule(lr){4-8}\cmidrule(lr){9-11}
& & DKTRec & DQN & AC & PPO & CSEAL & GEPKSD & ReAL & GenAL & IB-GRPO \\
\midrule
\multirow{3}{*}{Junyi}
 & 20 & -0.2599 & -0.2079 & 0.4950 & 0.4847 & 0.4106 & 0.3807 & \underline{0.5724} & 0.5692 & \textbf{0.7743} \\
 & 10 & -0.2604 & -0.1536 & 0.3135 & 0.3953 & 0.3574 & 0.3766 & 0.4001 & \underline{0.4243} & \textbf{0.6435} \\
 & 5  & -0.1467 & -0.0910 & 0.2490 & 0.3277 & \underline{0.4087} & 0.3304 & 0.2637 & 0.2145 & \textbf{0.6115} \\
\midrule
\multirow{3}{*}{ASSIST09}
 & 20 & -0.0284 & -0.2270 & 0.4500 & 0.4995 & 0.3484 & \underline{0.5837} & 0.4303 & 0.3665 & \textbf{0.5911} \\ 
 & 10 & -0.0576 & -0.1648 & 0.4314 & 0.4262 & 0.2865 & \underline{0.4737} & 0.3219 & 0.3234 & \textbf{0.5139} \\
 & 5  & -0.0365 & -0.1101 & \underline{0.3822} & 0.3201 & 0.1929 & 0.3647 & 0.2894 & 0.2504 & \textbf{0.4222} \\
\bottomrule
\end{tabular}
\end{table*}

 \begin{table}[!t]
\caption{Statistics of the datasets}
\label{t:data}
  \centering
  \begin{tabular}{l|c|c} 
    \toprule
    Statistics         & Junyi        & ASSIST09    \\
    \midrule
    Concepts & 835           & 167          \\
    Learners           & 525,061      & 4,217       \\
    Records  & 21,460,249         & 346,860          \\
    \bottomrule
  \end{tabular}
\end{table}

\subsubsection{Baselines}
To comprehensively evaluate model performance, we select the following methods as baselines for comparison:
\begin{itemize}
    \item \textbf{DKTRec:} A DKT-based greedy policy that recommends the concept whose predicted mastery is closest to 0.5.
\end{itemize}
\begin{itemize}
    \item \textbf{General RL} (DQN~\cite{mnih2013playing}, AC~\cite{konda1999actor}, PPO~\cite{ppo}): Generic baselines for sequential LPR without domain-specific design.
    \item \textbf{CSEAL~\cite{liu2019exploiting}:} Education-specific actor--critic with cognitive navigation for concept-level exploration.
    \item \textbf{GEPKSD~\cite{li2024privileged}:} Graph-based RL with privileged-teacher distillation.
\end{itemize}
\begin{itemize}
    \item \textbf{GenAL~\cite{lv2025genal}:} A general-purpose adaptive learning agent that directly utilizes the semantic understanding of LLMs to generate recommendations via prompting.
    \item \textbf{ReAL~\cite{lv2025real}:} The LLM-based education agent employs a "Retrieval-Reflection" mechanism to simulate a teacher's decision-making process.
\end{itemize}

\subsubsection{Implementation Details}
In our framework, we employ Qwen2.5-7B~\cite{qwen} as the backbone. All experiments were conducted on a server equipped with 8 NVIDIA A800 40-GB GPUs. Following the standard protocol~\cite{li2024privileged}, we split the datasets by student IDs to ensure no data leakage between training and testing. The simulator is trained on the training set to initialize student states, while the evaluation is performed on the held-out test set. More hyperparameter settings are shown in Appendix~A. The prompt templates used for SFT data and RL training are provided in Appendix~B. For the LLM-agent baselines ReAL and GenAL, we report the $E_p$ from their original papers~\cite{lv2025real,lv2025genal} in Table~\ref{tab1}; other baselines are evaluated under our unified KES simulator protocol.

\begin{figure}[!t]
    \centering
    \includegraphics[width=0.8\linewidth]{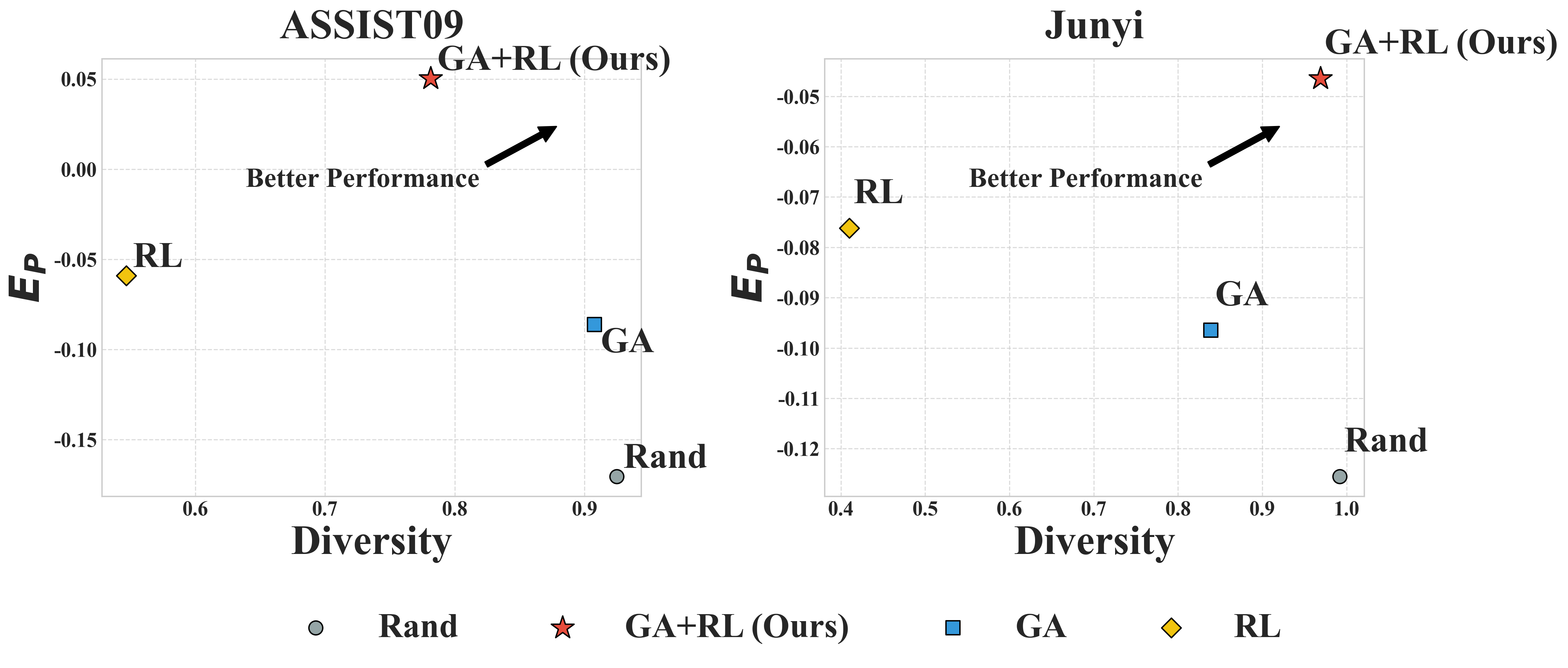}
    \caption{Pareto analysis of data synthesis strategies on ASSIST09 and Junyi datasets.}
    \label{fig2}
\end{figure}

\begin{figure}[!t]
    \centering
    \includegraphics[width=0.6\linewidth]{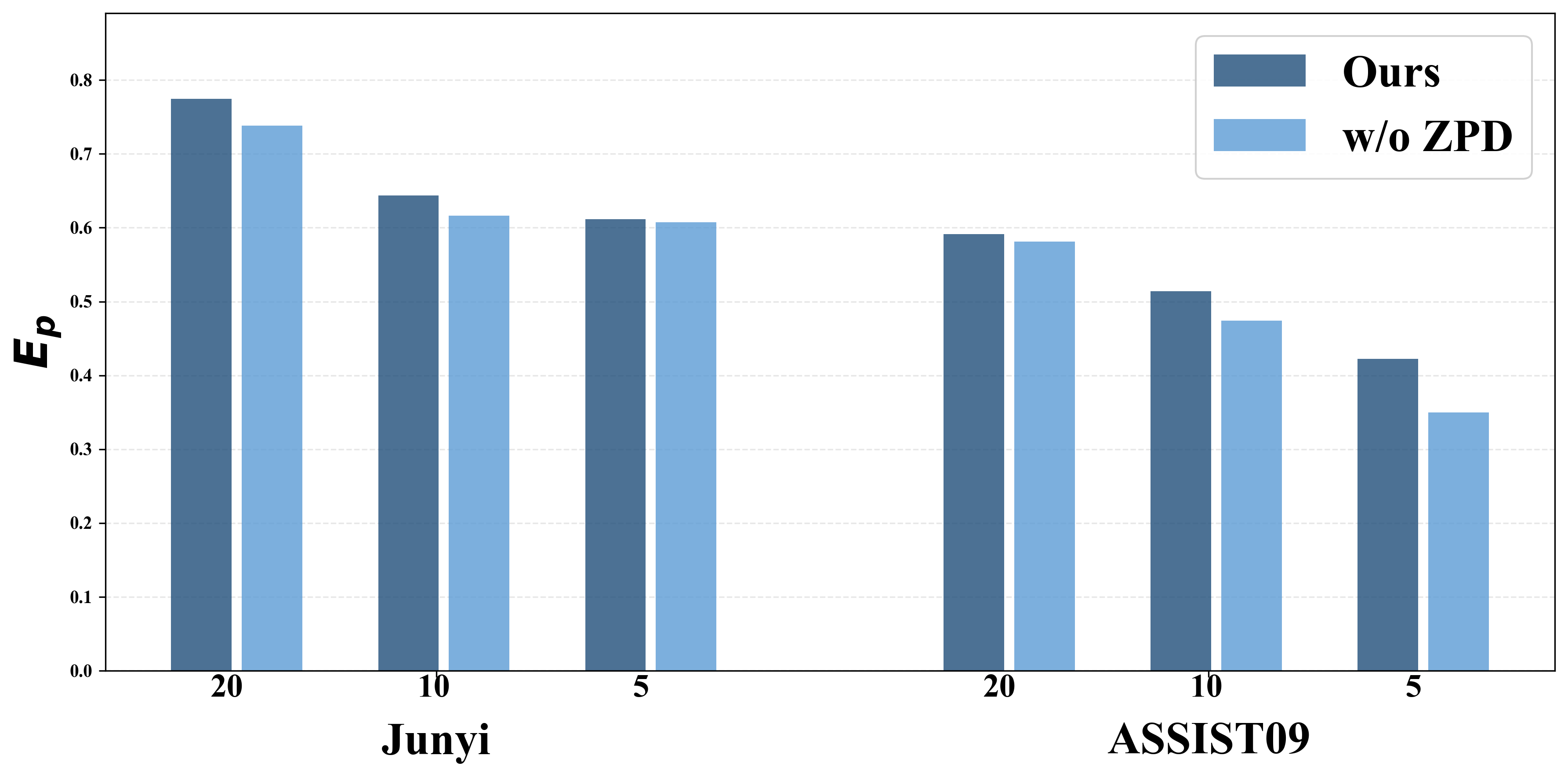}
    \caption{Ablation of the ZPD reward across path lengths ($L\in{5,10,20}$), evaluated by mean $E_p$ on ASSIST09 and Junyi.}
    \label{fig3}
 \end{figure}

\subsubsection{Diagnostic Metrics}
\label{sec:diagnostic_metrics}
To provide transparent and comparable diagnostics beyond training-time rewards, we report two additional metrics for analyzing generated paths.

For evaluation, we use a bounded score for length satisfaction to quantify how well the generated output length matches the target length $L_{\text{target}}$:
\begin{equation}
\mathrm{LenScore}(\pi)
=
\max\left\{0,\,1-\frac{\left||\pi|-L_{\text{target}}\right|}{L_{\text{target}}}\right\}.
\end{equation}
This score maps length deviations to $[0,1]$. We emphasize that $\mathrm{LenScore}$ is an evaluation metric, while the training objective uses the length reward $R_{Len}$ (Section~3.1) with tolerance and penalty parameters to shape the optimization landscape.

We report trajectory-level diversity as the within-path unique-concept ratio on fixed-length rollouts:
\begin{equation}
\mathrm{Div}_{\text{path}}(\pi)=\frac{|\mathrm{uniq}(\pi)|}{L}.
\end{equation}
This simple diagnostic reflects distinct concept coverage within a single path. It differs from the training-time diversity reward $D_{Div}$, which measures prompt-grouped inter-sample diversity via $n$-gram Jaccard dissimilarity to encourage exploration and mitigate mode collapse. In short, we use $D_{Div}$ as an optimization regularizer during policy learning, while $\mathrm{Div}_{\text{path}}$ serves as a transparent diagnostic for analyzing the resulting paths.

\subsection{Overall Performance Comparison (RQ1)}

\begin{figure}[!t]
    \centering
    \includegraphics[width=0.7\linewidth]{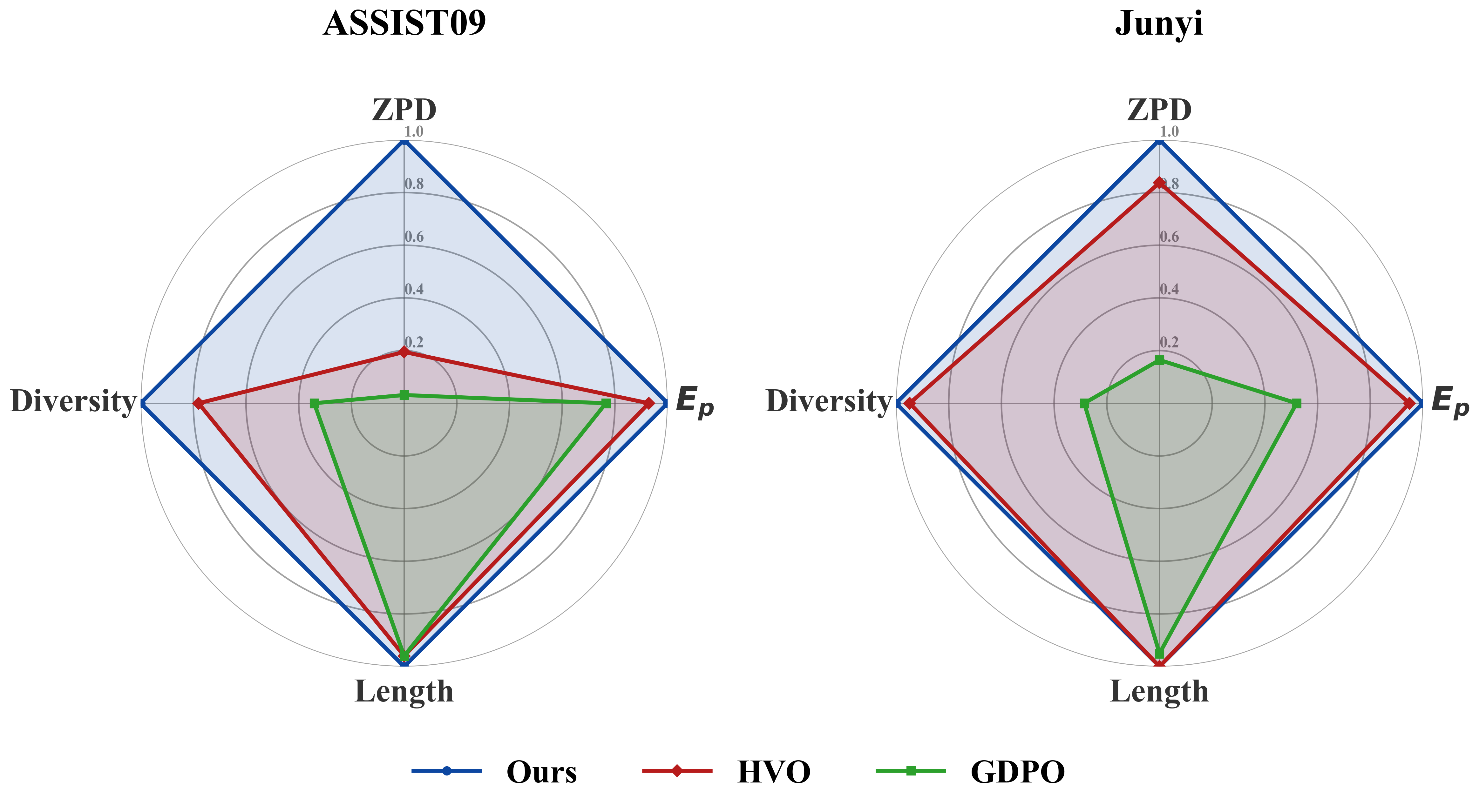}
    \caption{The axes correspond to Learning Effect ($E_p$), ZPD alignment score ($S_{ZPD}$), trajectory-level diversity ($\mathrm{Div}_{\text{path}}$), and length satisfaction score ($\mathrm{LenScore}$).
}
    \label{fig4}
 \end{figure}

From the experimental results in Table~\ref{tab1}, we observe that IB-GRPO achieves the best performance across all datasets and recommendation lengths. Compared to the RL method GEPKSD, IB-GRPO benefits from a demonstration-based warm start and indicator-based multi-objective advantage estimation. Compared to the prompting-based LLM agent ReAL, IB-GRPO performs explicit policy optimization with trajectory-level feedback (including delayed learning outcomes and ZPD-related signals), helping the policy maintain pedagogically aligned difficulty throughout long horizons and reducing myopic behavior or late-stage difficulty drift.

\subsection{Pareto Analysis of Data Generation (RQ2)}

To assess the hybrid data construction mechanism, we compare SFT policies trained from four data sources in terms of $E_p$ and the diversity diagnostic $\mathrm{Div}_{\text{path}}$. Figure~\ref{fig2} shows that different sources yield distinct $E_p$--diversity profiles: Rand (uniform random paths) achieves high diversity but very low $E_p$, indicating that diversity without pedagogical structure is ineffective. RL-only collapses to a narrow set of low-diversity paths, while GA-only explores more broadly but often remains in moderate-$E_p$ regions. In contrast, GA+RL shifts the distribution toward higher $E_p$ while retaining substantial diversity, yielding a better empirical Pareto envelope in the $E_p$--Diversity plane and providing a stronger warm start for subsequent IB-GRPO training.

\subsection{Component Impact Study (RQ3)}

We further investigate the contributions of individual components. Figure~\ref{fig3} reveals that while the variant without ZPD achieves comparable performance at short horizons ($L=5$) on Junyi, our method consistently outperforms it as the path length extends to $L=20$ across both datasets. This confirms that the ZPD reward is essential for preventing myopic behaviors and ensuring sustainable learning effects in long-horizon planning.
\begin{figure}[t!]
    \centering
    \includegraphics[width=0.7\linewidth]{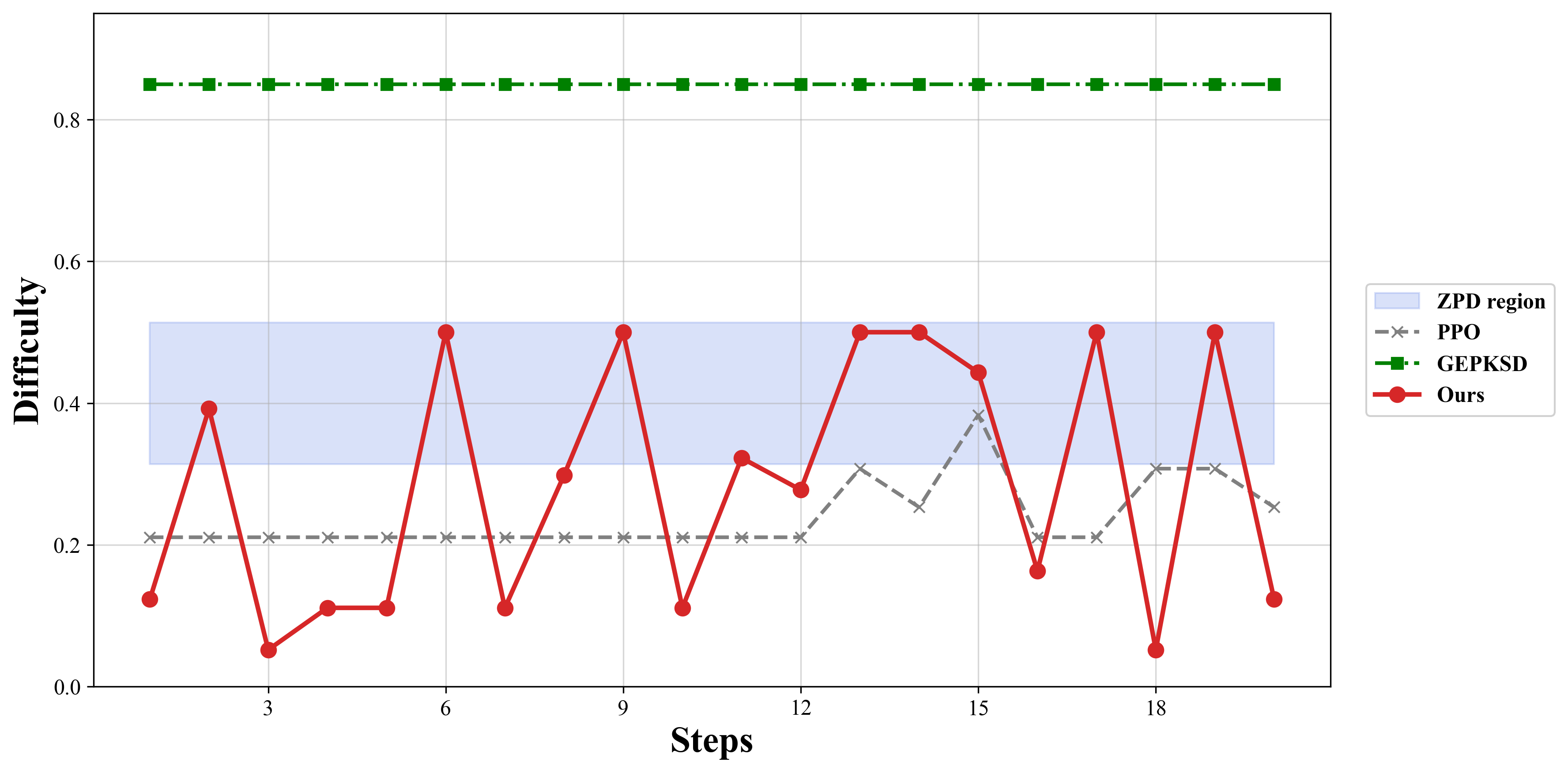}
    \caption{Trajectory comparison of recommended concept difficulty relative to the ZPD band. The blue area visualizes the ZPD region (centered on $z(a)$ with $\pm$$\sigma$ band).}
    \label{fig:trace}
\end{figure}

Figure~\ref{fig4} compares IB-GRPO with recent baselines, including HVO~\cite{song2025balancing} and GDPO~\cite{liu2026gdpo}. For fairness, all methods share the same backbone, training pipeline, and hyperparameters, and differ only in how the multi-objective reward vector is converted into optimization signals. For visualization, each axis is independently normalized to $[0,1]$ across methods on the same dataset. 
Across both datasets, IB-GRPO yields a better overall trade-off among $E_p$, $S_{ZPD}$, trajectory-level diversity, and length satisfaction. HVO and GDPO often meet the length constraint but show lower $S_{ZPD}$ and diversity, suggesting less balanced trade-offs. 

\subsection{Case Study: Visualization of Difficulty Trace (RQ4)}

Figure~\ref{fig:trace} visualizes the difficulty trajectories of recommended paths for a sampled student (ID: 931) under different policies, relative to the empirically defined ZPD band (blue shaded area, centered on $z(a)$ with a $\pm\sigma$ margin). The PPO baseline shows large fluctuations in difficulty and frequently departs from the ZPD band. GEPKSD tends to remain in lower-difficulty ranges for extended segments, which may under-challenge the learner. In contrast, our policy (red line) stays closer to the ZPD band throughout the horizon, indicating better difficulty alignment in this illustrative case. We additionally include a supplementary figure in the Appendix that prints the step-wise knowledge-point names of the recommended path for the sampled student.

\section{Conclusion}
We propose IB-GRPO, an indicator-guided alignment approach for LLM-based LPR, which optimizes learning effect while encouraging ZPD-consistent difficulty scheduling and satisfying operational constraints on length and path diversity. IB-GRPO (i) constructs hybrid expert demonstrations via GA search and offline RL teachers for reliable warm-start, (ii) instantiates ZPD as a computable reward for within-session difficulty alignment, and (iii) introduces an $I_{\epsilon+}$-indicator-based intra-group advantage to avoid manual scalarization and better capture Pareto trade-offs. On both ASSIST09 and Junyi, IB-GRPO consistently achieves the Pareto-balanced performance across multiple horizons, improving learning effect while preserving ZPD-aligned difficulty and trajectory-level diversity.
\bibliographystyle{named}
\bibliography{main}

\end{document}


\maketitle

\appendix  

\section{Hyperparameters} 

\begin{table}[H]
    \centering
    \caption{Training setups and hyperparameters.}
    \label{tab:hyperparameters}
    \begin{tabular}{lc} 
        \toprule
        \textbf{Parameter} & \textbf{Value} \\
        \midrule
        \multicolumn{2}{l}{\textit{Supervised Fine-Tuning (SFT)}} \\
        \quad Learning Rate & $5.0 \times 10^{-6}$ \\
        \quad Number of Epochs & 10 \\
        \quad Warmup Ratio & 0.1 \\
        \quad Temperature & 0.5 \\
        \midrule
        \multicolumn{2}{l}{\textit{IB-GRPO}} \\ 
        \quad Learning Rate & $1 \times 10^{-6}$ \\
        \quad Batch Size & 16 \\
        \quad Number of Epochs & 5 \\   
        \quad Group Size & 8 \\
        \quad Warmup Steps & 10 \\
        \quad Temperature & 1 \\     
        \midrule
        \multicolumn{2}{l}{\textit{Genetic Algorithm (GA)}} \\
        \quad Population Size & 10 \\
        \quad Crossover Probability & 1.0 \\
        \quad Mutation Probability & 0.01 \\
        \quad Number of Generations & 500 \\
        \midrule
        \multicolumn{2}{l}{\textit{Environment \& Reward Settings}} \\ 
        \quad $I_{\epsilon+}$ Scaling Factor $\kappa$ & 0.05 \\
        \quad Length Constraint Tolerance $\tau$ & 1 \\
        \quad Length Constraint Penalty $\lambda$ & 0.1 \\
        \quad ZPD Reward Distribution Width $\sigma$ & 0.1 \\
        \quad n-gram size $n$ (for $D_{Div}$) & 2 \\
        \bottomrule
    \end{tabular}
\end{table}

\begin{figure}
    \centering
    \includegraphics[width=1.1\linewidth]{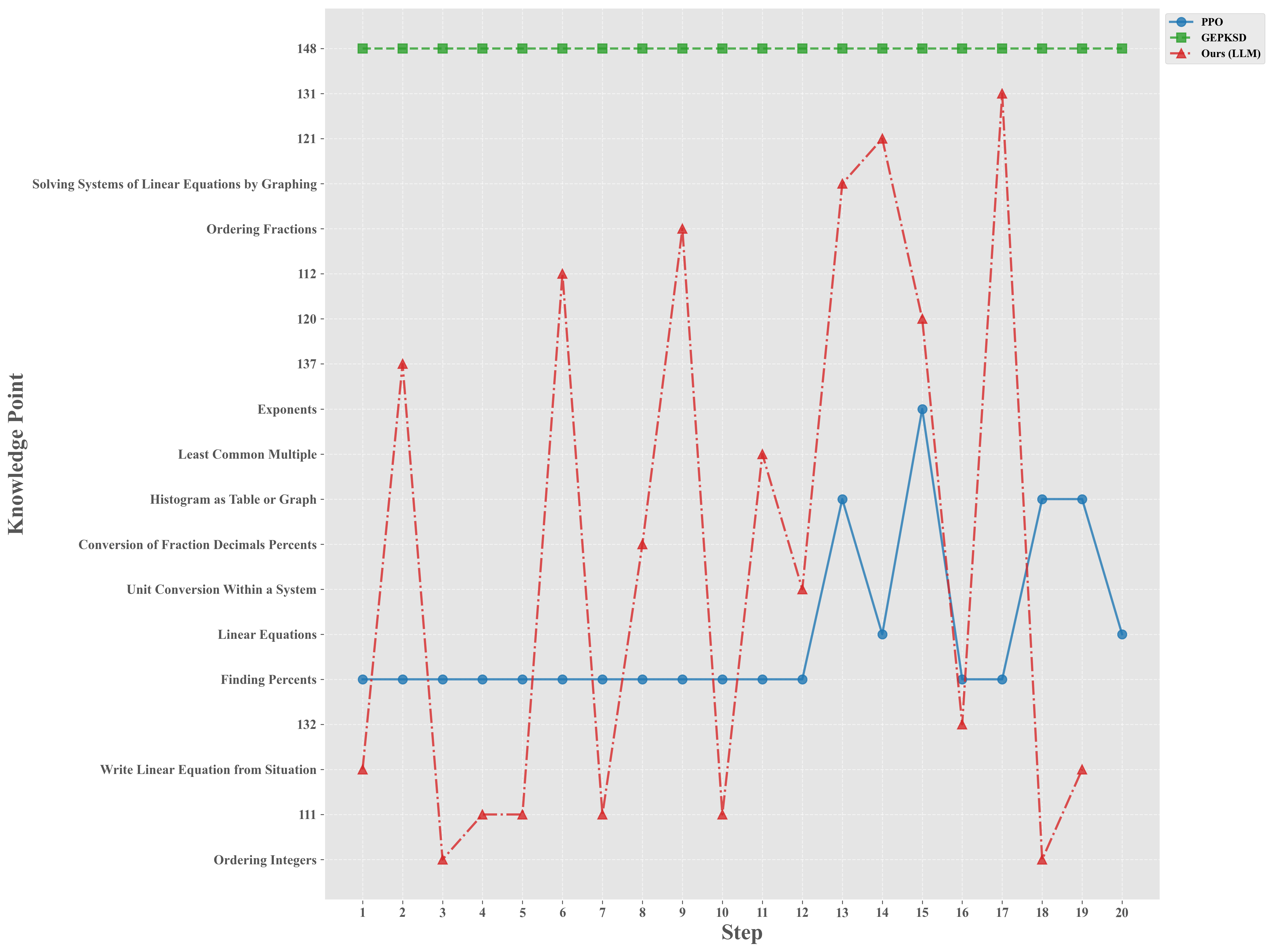}
    \caption{Visualization of step-wise knowledge-point names for the sampled student under different policies.}
    \label{fig:placeholder}
\end{figure}

\begin{figure}
    \centering
    \includegraphics[width=1\linewidth]{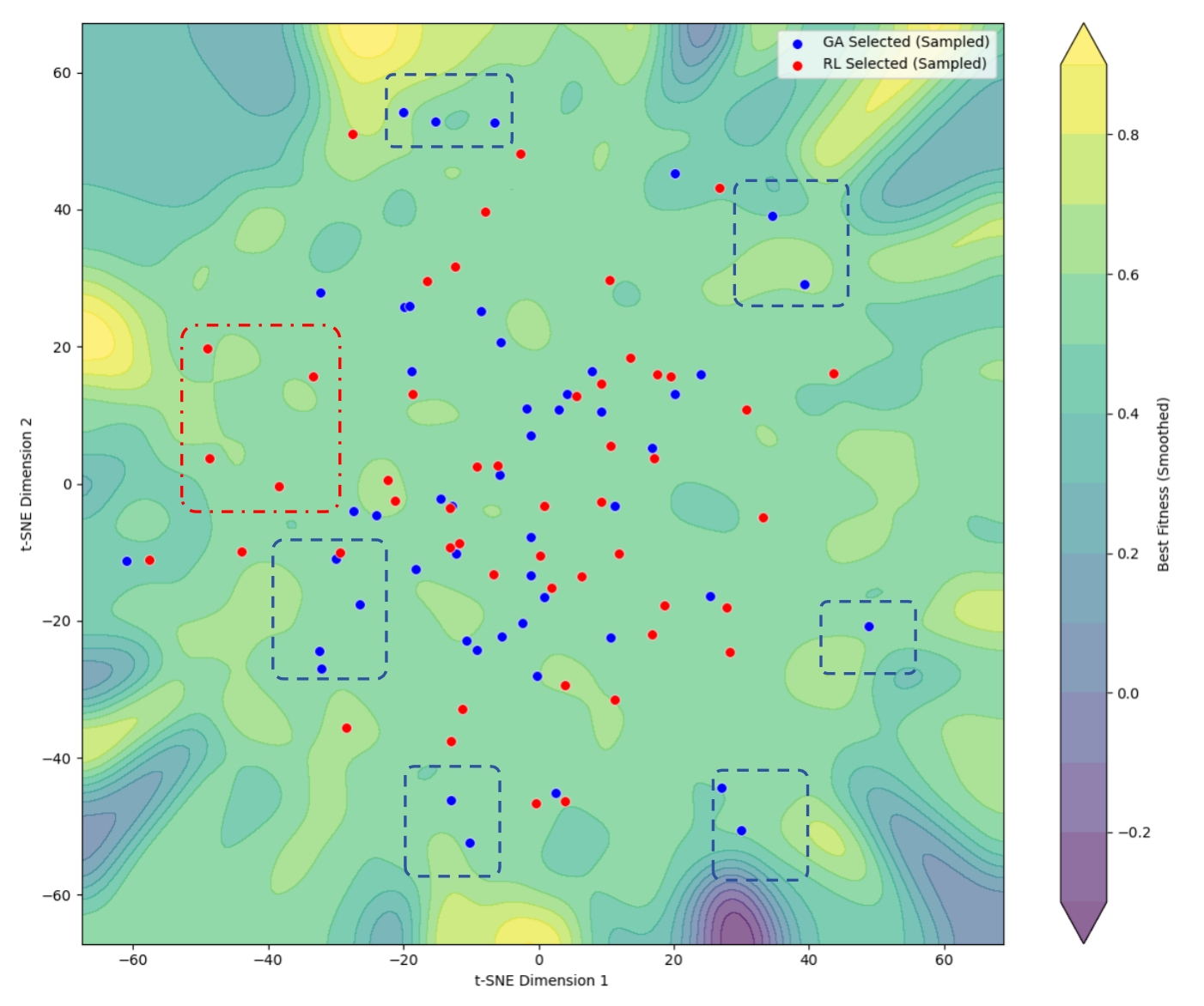}
    \caption{Visualization of SFT data generated by GA and RL}
    \label{fig:placeholder}
\end{figure}

\section{Step-wise Knowledge-Point Names}
\label{app:kp-names}

Figure~\ref{fig:placeholder} shows the knowledge-point name selected at each step for the sampled student under different policies. The y-axis represents the ordinal rank of exercise difficulty (lower is easier).

\section{Prompt Design for Learning Path Recommendation} 

To ensure the LLM effectively interprets educational context and adheres to pedagogical constraints, we designed a structured prompt template. As detailed in Table~\ref{tab:prompt_detailed}, this structure is consistently applied across both the SFT and RL stages, as well as across all experimental datasets (ASSIST09 and Junyi). 

\section{Supplementary information} 
We visualize the distribution of the learning paths generated by the Genetic Algorithm (GA) and Reinforcement Learning (RL)  in Figure~\ref{fig:placeholder}. As can be seen from the figure, these two methods exhibit distinct yet highly complementary characteristics near the yellow areas  with high fitness values. To visualize the dataset, we generated semantic embeddings for each exercise, subsequently projecting these high-dimensional representations into a two-dimensional space via t-SNE.


\begin{table*}[t]
\centering
\caption{LPR Prompt Template and Output Format. Blue highlights contextual cues (decision-relevant signals), while red denotes hard constraints and strict output rules.}
\label{tab:prompt_detailed}
\small
\begin{tabular}{p{2.5cm}p{12cm}}
\toprule
\textbf{Component} & \textbf{Content} \\
\midrule
\textbf{[System Message]} & You are an \textcolor{blue} {experienced teaching professional} tasked with recommending learning paths based on students' \textcolor{blue} {historical mastery and learning targets}. \\
 & You are an intelligent recommendation engine in an educational system. Based on the information below, recommend N knowledge concept IDs that the student should focus on next. \\
\midrule
\textbf{[User Input]} & \textbf{1. Target learning points for the student:} \\
 & Format: JSON object mapping knowledge concept IDs to names \\
 & Example: \{"96": "divisibility\_0.5", "98": "least\_common\_multiple"\} \\
 & \\
 & \textbf{2. The student's last 50 question records (history logs):} \\
 & Format: Array of objects with fields: id, name, correct \\
 & Example: [\{id: 432, name: "subtraction\_2", correct: 1\}, \{id: 93, name: "addition\_4", correct: 1\}, ...] \\
 & Contains: Knowledge concept ID, \textcolor{blue} {concept name, and correctness} (0 or 1) \\
 & \\
 & \textbf{3. Candidate knowledge concepts:} \\
 & Total of $\mathcal{C}$ available knowledge concepts \\
 & Selection \textcolor{red} {must be from this candidate set} \\
\midrule
\textbf{[Constraint]} & The path length should be \textcolor{red} {N steps} (N = 5, 10, or 20) \\
 & Only select knowledge concept IDs from the \textcolor{red} {$|\mathcal{C}|$ candidate knowledge concepts} \\
 & Your selection should be relevant to the student's target learning points and \textcolor{blue} {informed by their recent activity} \\
 & Ensure diversity and progression in the recommendations where appropriate \\
\midrule
\textbf{[Output]} & Provide the recommended sequence of knowledge concept \textcolor{red} {IDs only} \\
 & Output format: JSON array of N integers \\
 & Example: [182, 42, 31, 23, ..., 123] \\
 & \textcolor{red} {No comments or explanations} \\
 & Strict output format: [id1, id2, ..., idN] \ \\
\bottomrule
\end{tabular}
\end{table*}